\documentclass{article}
\usepackage{spconf,amsmath,graphicx}
\usepackage{float}
\usepackage{stfloats}
\usepackage{subcaption}
\usepackage{url}
\usepackage{array}
\usepackage[table]{xcolor}
\usepackage{float}
\usepackage{caption}
\usepackage{booktabs}
\usepackage{graphicx}
\usepackage{romannum}
\usepackage{makecell}

\usepackage{enumitem}
\setlist{nosep, leftmargin=14pt}

\usepackage{mwe} 

\usepackage{xcolor}

\title{SLD: Segmentation-Based Landmark Detection for Spinal Ligaments}
\name{Lara Blomenkamp$^1$ \qquad Ivanna Kramer$^1$ \qquad Sabine Bauer$^2$ \qquad Theresa Schöche$^3$}

\address{$^1$ Institute for Computational Visualistics, University of Koblenz, Germany \\
    $^2$ University Sports, University of Koblenz, Germany \\
    $^3$ Institute of Anatomy, University Medical Center of Johannes Gutenberg University Mainz, Germany
    }

\begin{document}
%
\maketitle
\begin{abstract}
In biomechanical modeling, the representation of ligament attachments is crucial for a realistic simulation of the forces acting between the vertebrae. These forces are typically modeled as vectors connecting ligament landmarks on adjacent vertebrae, making precise identification of these landmarks a key requirement for constructing reliable spine models.
Existing automated detection methods are either limited to specific spinal regions or lack sufficient accuracy.
This work presents a novel approach for detecting spinal ligament landmarks, which first performs shape-based segmentation of 3D vertebrae and subsequently applies domain-specific rules to identify different types of attachment points. The proposed method outperforms existing approaches by achieving high accuracy and demonstrating strong generalization across all spinal regions. Validation on two independent spinal datasets from multiple patients yielded a mean absolute error (MAE) of 0.7 mm and a root mean square error (RMSE) of 1.1 mm.\footnote{This work is accepted at IEEE International Symposium on Biomedical Imaging (ISBI) 2026} 
\end{abstract}

\begin{figure*}[!hb]
  \centering
  \includegraphics[width=0.9\textwidth]{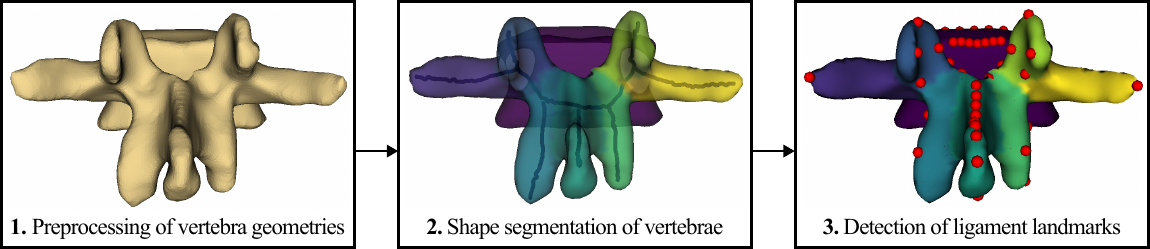}
  \caption{Steps of the proposed SLD method for detecting spinal ligament landmarks.}
  \label{fig:method-steps}
\end{figure*}

\section{Introduction}
\label{sec:intro}

The human spine is a complex structure composed of a series of vertebral bones, separated by intervertebral discs, and interconnected by ligaments.
Through biomechanical modeling of the spine, the load distribution and biomechanical behavior of these spinal components can be determined and analyzed. 
In an accurate, patient-specific biomechanical simulation model, spinal structures - including vertebral bodies, intervertebral discs, and ligaments - must reflect individual anthropometric characteristics that are essential for realistic patient-specific modeling.
Accordingly, several sets of points define the anatomical attachment sites of the spinal ligaments on the vertebral surfaces (see Fig.~\ref{fig:method-steps} (3)). Within the biomechanical model, these attachment points are connected to compute force vectors representing ligament tension.
These simulations enable a wide range of applications, including research on anatomical variations, surgical procedures, devices, and implants, or the investigation and prediction of biomechanical properties during physical activities  \cite{Campbell.2015}.
The accurate detection of spinal ligament landmarks is therefore essential for constructing biomechanical models used in simulations of spinal forces.
Identifying soft tissues such as ligaments in medical images is challenging due to noise and low-resolution artifacts, as well as the complexity and variability of the musculoskeletal system.
Additionally, direct segmentation methods are often noise-sensitive, unreliable, and inaccurate, making the task difficult to automate \cite{Assassi2009FMT}. 
To enhance the workflow of creating simulation models of the spine, it is necessary to improve the accuracy and efficiency.
Automating the identification of ligament landmarks on vertebrae is a challenging task due to their complex morphology, which includes fine and intricate anatomical features such as various bony processes. Moreover, vertebral morphology exhibits substantial regional variation across the cervical, thoracic, and lumbar spine, which may be further complicated by pathological alterations. 
Therefore, existing studies often target only the cervical spine \cite{AlDhamari2019ADO} or the lumbar spine \cite{Campbell.2015, lerchl2022validation}.
The identification of anatomical landmarks in medical data has been widely investigated, primarily using registration-based techniques \cite{Bromiley2014SemiautomaticLP, young2015, Porto2021ALP}, while methods specific to spinal ligament landmark detection rely either on registration \cite{AlDhamari2019ADO, Kramer.2024} or on direct geometric and shape analysis \cite{Campbell.2015, lerchl2022validation}.

However, current approaches have been observed to yield limited accuracy in identifying anatomical landmarks on the complex morphology of vertebrae, particularly in regions exhibiting high structural variability or irregular shapes.
This highlights the need for a method that can reliably and accurately identify ligament landmarks across all spinal regions.
Our contribution is threefold: We propose a novel method for detecting spinal ligament landmarks, we publish a dataset of annotated landmarks, validated by medical experts, and we provide the implementation as a publicly available and ready-to-use plugin \footnote{\raggedright\url{https://github.com/VisSim-UniKO/SpineToolkit}}.

\section{Materials and Methods}
\label{sec:methods}


The goal of the proposed method is the detection of ligament landmarks on segmented anatomical regions of 3D vertebra geometries.
First, the geometries are prepared for the subsequent segmentation and landmark detection methods, by applying remeshing and smoothing techniques (see Fig.~\ref{fig:method-steps} (1.)).
This ensures the required quality and usability of the mesh data, which may stem from different datasets with different properties.
Additionally, a first analysis of the geometry is conducted to extract certain features of the vertebrae, such as their dimensions and orientation, which are used as parameters for subsequent steps.
In the second step, the vertebra geometries are segmented into their anatomical components (see Fig.~\ref{fig:method-steps} (2.)). 
Finally, the ligament landmarks are detected on the segmented geometries (see Fig.~\ref{fig:method-steps} (3.)).

\subsection{Shape Segmentation of 3D Vertebrae}
\begin{figure}[htb]
    \centering
    \begin{subfigure}[b]{0.49\linewidth}
        \centering
        \includegraphics[width=\linewidth]{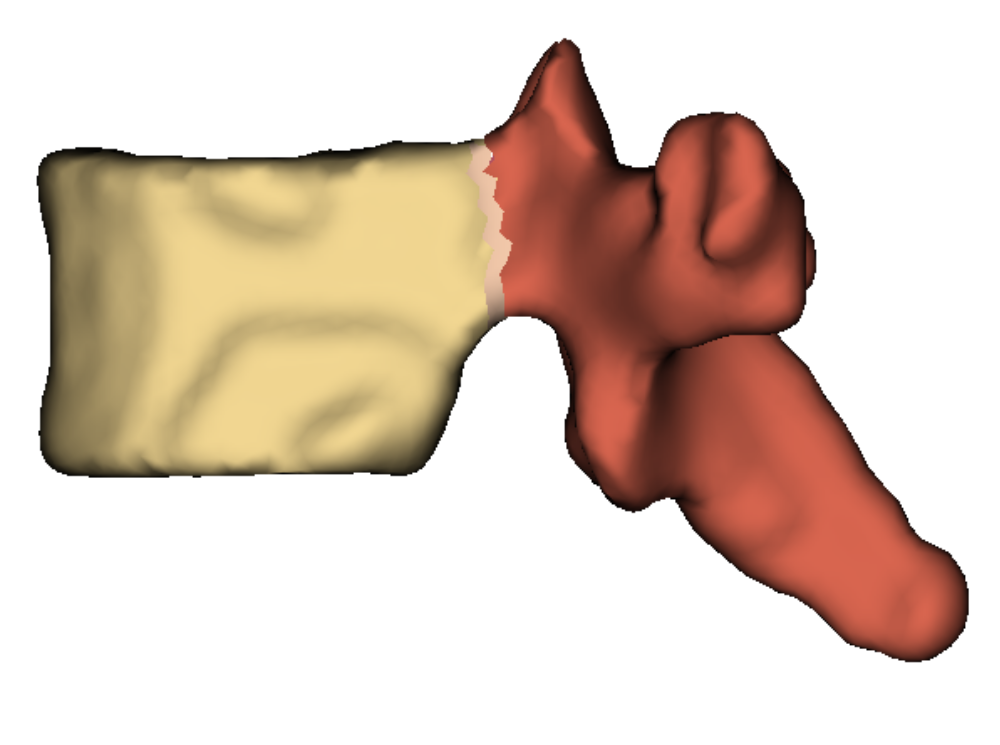}
        \caption*{\parbox{\linewidth}{\centering (a) Extraction of vertebral\\body and vertebral arch}}
    \end{subfigure}
    \hfill
    \begin{subfigure}[b]{0.49\linewidth}
        \centering
        \includegraphics[width=\linewidth]{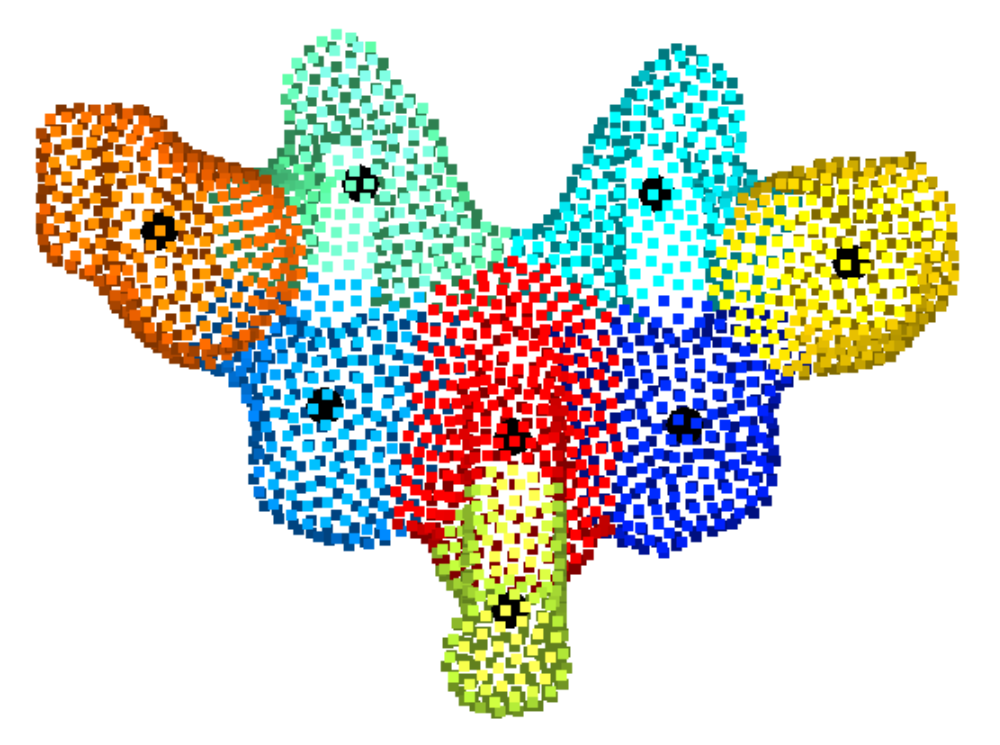}
        \caption*{\parbox{\linewidth}{\centering (b) Clustering for identification\\of geometrical extremes}}
    \end{subfigure}

    \vspace{1em}

    \begin{subfigure}[b]{0.49\linewidth}
        \centering
        \includegraphics[width=\linewidth]{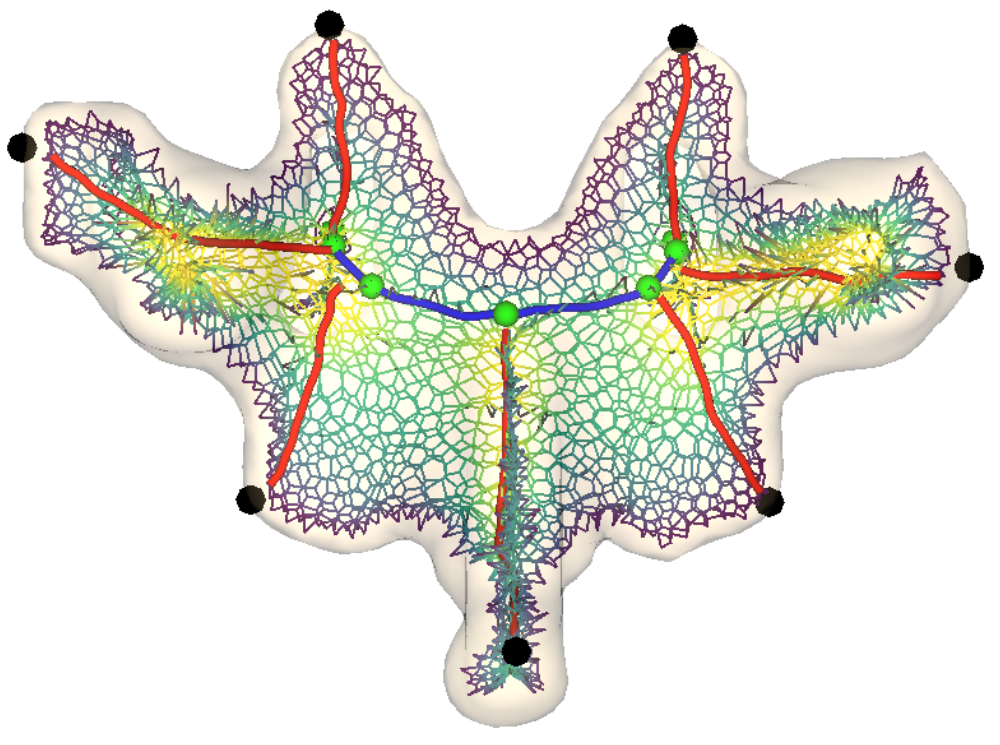}
        \caption*{\parbox{\linewidth}{\centering (c) Skeletonization}}
    \end{subfigure}
    \hfill
    \begin{subfigure}[b]{0.49\linewidth}
        \centering
        \includegraphics[width=\linewidth]{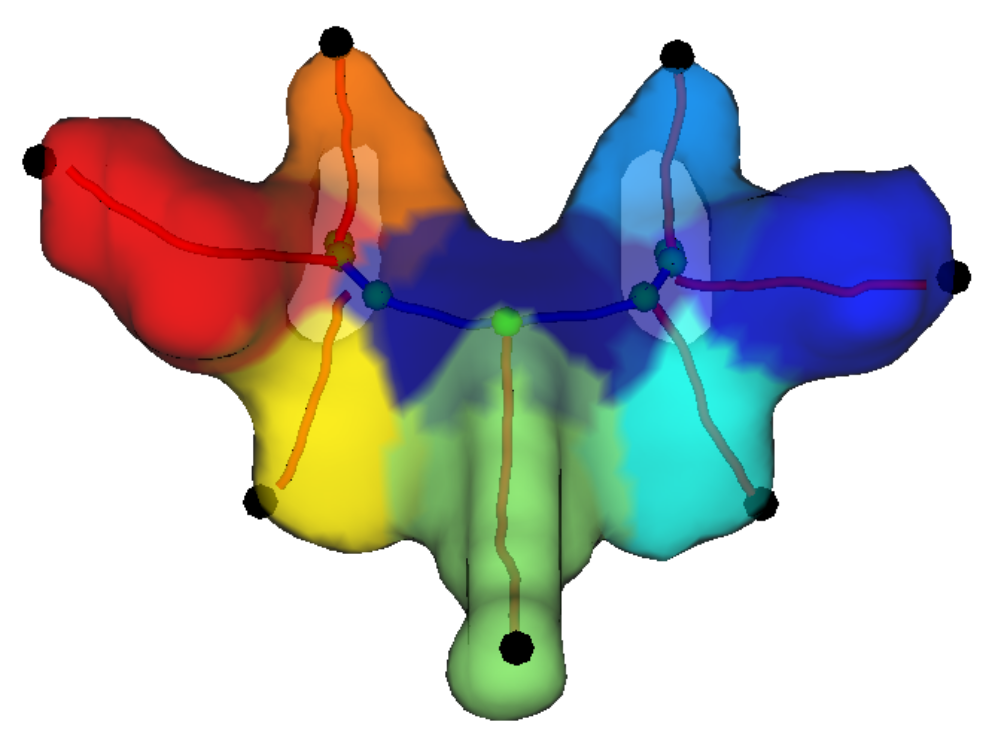}
        \caption*{\parbox{\linewidth}{\centering (d) Segmentation}}
    \end{subfigure}

    \caption{Steps of the skeletonization-based method for semantic segmentation of vertebrae \cite{Blomenkamp.2025}.}
    \label{fig:pipeline-segmentation}
\end{figure}

The anatomical regions of the vertebrae are divided with a skeletonization-based approach for shape segmentation \cite{Blomenkamp.2025}, as displayed in Fig.~\ref{fig:pipeline-segmentation}.

First, the vertices of the vertebra mesh are seperated into the vertebral body and the vertebral arch (see Fig.~\ref{fig:pipeline-segmentation} (a)).
Subsequently, the vertebral arch is further segmented into its components: the lamina, the spinous process, the left and right transverse processes, two superior articular processes, and two inferior articular processes.
This is accomplished by clustering the vertices into initial segments and identifying geometrical extreme points for the segments (see Fig.~\ref{fig:pipeline-segmentation} (b)). Then, a skeleton curve is calculated for each required segment to represent the general shape of the anatomy (see Fig.~\ref{fig:pipeline-segmentation} (c)). To obtain the final segmentation (see Fig.~\ref{fig:pipeline-segmentation} (d)), all vertices are classified based on their distance to the skeleton curves. This approach is explained in more detail in \cite{Blomenkamp.2025}.

\subsection{Detection of Spinal Ligament Landmarks}

\begin{figure*}[!h]
    \centering
    \begin{subfigure}[b]{0.21\textwidth}
        \centering
        \includegraphics[width=\linewidth]{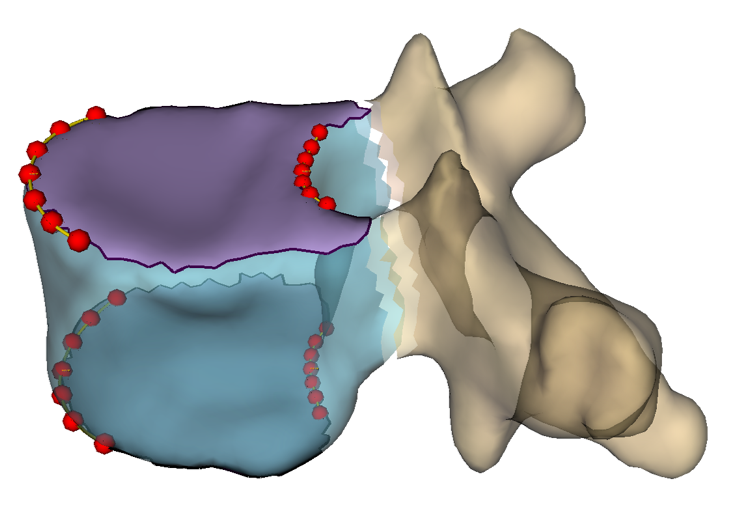}
        \caption*{\parbox{\linewidth}{\centering (a) ALL and PLL}}
    \end{subfigure}
    \begin{subfigure}[b]{0.18\linewidth}
        \centering
        \includegraphics[width=\linewidth]{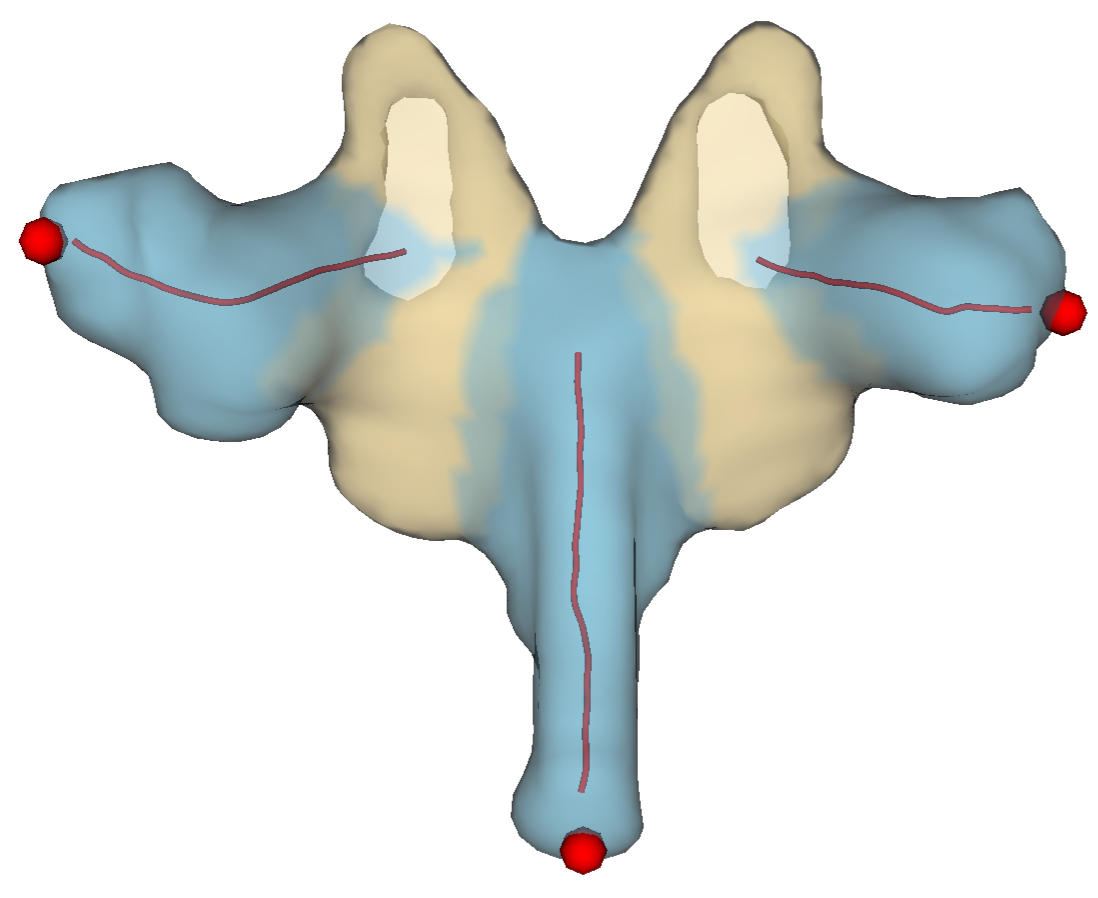}
        \caption*{\parbox{\linewidth}{\centering (b) ITL and SSL}}
    \end{subfigure}
    \hfill
    \begin{subfigure}[b]{0.16\linewidth}
        \centering
        \includegraphics[width=\linewidth]{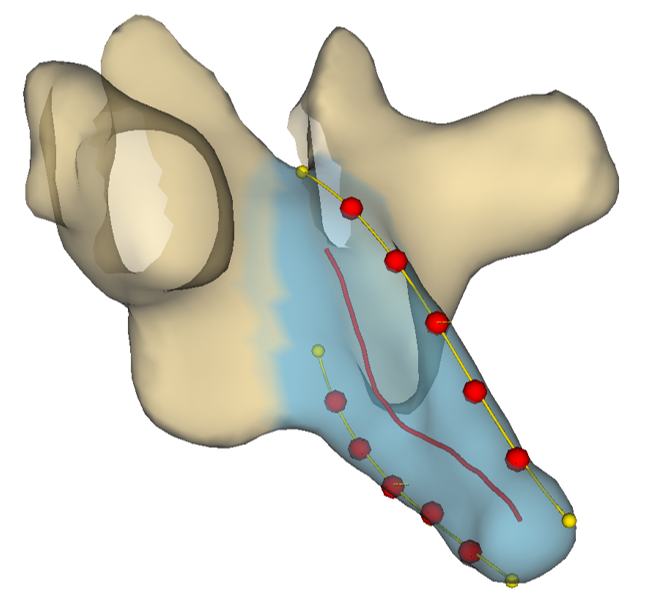}
        \caption*{\parbox{\linewidth}{\centering (c) ISL}}
    \end{subfigure}
    \hfill
    \begin{subfigure}[b]{0.18\linewidth}
        \centering
        \includegraphics[width=\linewidth]{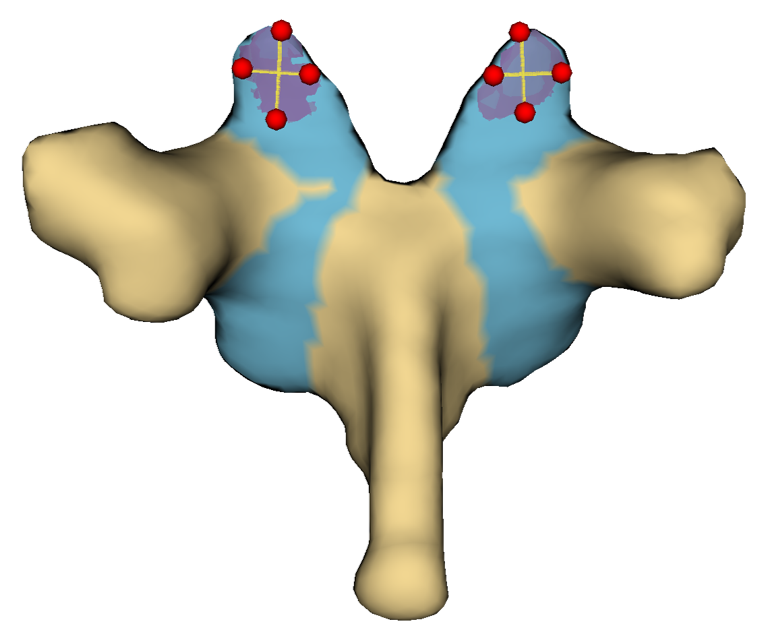}
        \caption*{\parbox{\linewidth}{\centering (d) CL}}
    \end{subfigure}
    \hfill
    \begin{subfigure}[b]{0.18\linewidth}
        \centering
        \includegraphics[width=\linewidth]{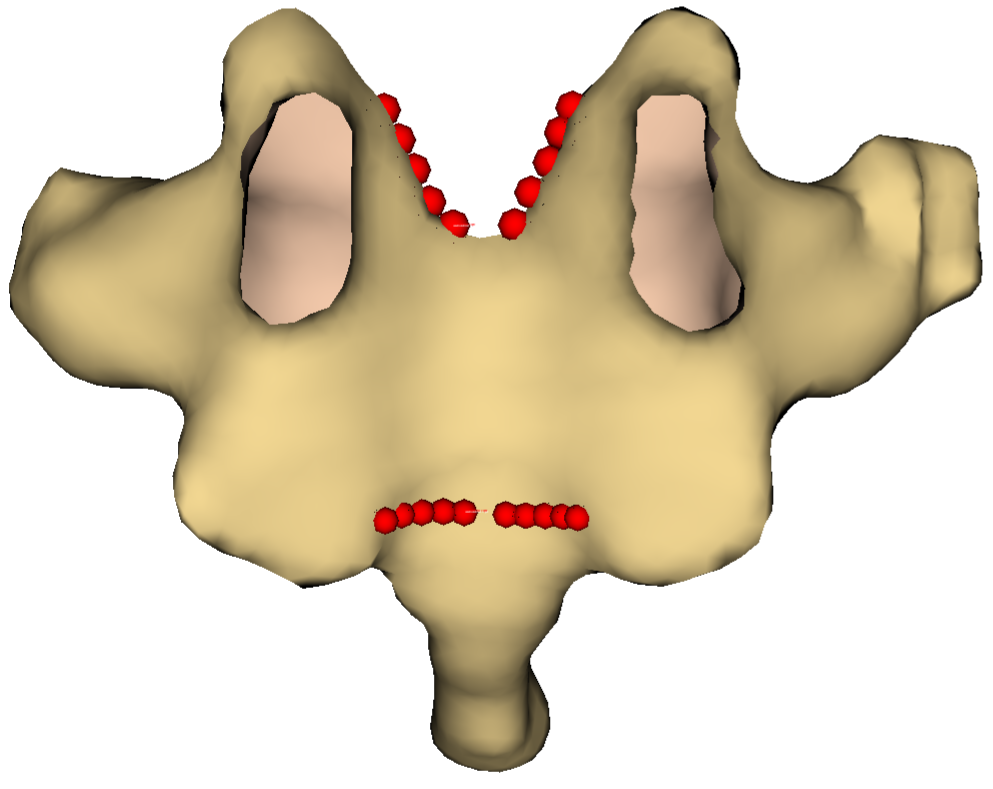}
        \caption*{\parbox{\linewidth}{\centering (e) LF}}
    \end{subfigure}

    \caption{Detection of the spinal ligament landmarks (red) on the corresponding vertebral regions (blue).}
    \label{fig:detection}
\end{figure*}

The localization of the ligament landmarks is performed on the corresponding vertebra regions of the previously segmented geometry.
They are detected using the following methods for fitting them to the surface of the geometry:

\subsubsection{ALL and PLL}
The \emph{anterior and posterior longitudinal ligaments (ALL and PLL)} connect the anterior and posterior edges of the vertebral bodies. Their ligament landmarks (see Fig.~\ref{fig:detection} (a)) are fitted to the boundary of the endplates surfaces, which are defined by selecting polygons with high similarity between the longitudinal axis $a_L$ and the polygons surface normal $n$. The polygons of the superior endplate are thereby extracted with the following condition:
\begin{equation}
    cos(\theta) \leq
    n
    \cdot
    a_L,
\end{equation}
and respectively for the inferior endplate, with the negative of the longitudinal axis. The landmarks are then equidistantly fitted to the anterior (ALL) and posterior (PLL) edge of both endplate surfaces. 

\subsubsection{ITL and SSL}
The \emph{intertransverse ligaments (ITL)} are represented by one landmark on each transverse process, determined as geometric extremes along the local orientation of the process. The \emph{supraspinous ligament (SSL)} is represented by one landmark at the tip of the spinous process. The ITL and SSL landmarks are detected on the surface of the corresponding process segment by extending their skeleton curves and then calculating the intersection with the surface (see Fig.~\ref{fig:detection} (b)).

\subsubsection{ISL}
The \emph{interspinous ligaments (ISL)} connect the spinous processes of adjacent vertebrae. In the biomechanical model, they are represented with two landmark groups, on the superior and the inferior edge of the process. The landmarks are fitted by projecting the corresponding skeleton curve upwards and downwards (see Fig.~\ref{fig:detection} (c)).
More precisely, sample points $c_i$ of the skeleton curve $\mathcal{C}_p$ are projected along a local longitudinal axis $a_L$ of the spinous process. 
The projecting lines $\Lambda$ upwards and downwards along $a_L$ are calculated as follows:

\begin{equation}
    \Lambda = \left\{ \lambda_i \mid \lambda_i(t) = c_i + t \cdot a_L, \; c_i \in \mathcal{C}_p, \; t \in {\rm I\!R} \right\}.
\end{equation}

The intersection points of the spinous process surface and the projecting lines $\Lambda$ are used to fit a superior and an inferior ISL curve, which are then equidistantly resampled to define the ligament landmarks.

\subsubsection{CL}
The \emph{capsular ligaments (CL)} are represented by four landmarks on each of the four facet surfaces. (see Fig.~\ref{fig:detection},~(d)). The facet surfaces of the articular processes are extracted by localizing the closest surface to the adjacent articular process, similar to the approach in \cite{Paccini.2023}.
For each pair of adjacent articular processes segments $A_1$ and $A_2$, the facet surface $J$ is defined as the set of vertices $x_i$ in $A_1$ with the minimal euclidean distances to the vertices $y_i$ in $A_2$:

\begin{equation}
    J = \left\{ x_i \in A_1 \mid x_i = \arg\min_{x \in A_1} \vert\vert x - y \vert\vert, \; \forall y \in A_2 \right\}.
\end{equation}

For each facet, two intersections are calculated from the surface and local transverse and sagittal planes. Their geometric extreme points then define the ligament landmarks. 

\subsubsection{LF}
The ligament landmarks of the \emph{ligamentum flavum (LF)} are fitted to the shortest paths on the vertebra surface between previously calculated CL landmarks (see Fig.~\ref{fig:detection} (e)). The paths are calculated as geodesic paths with Dijkstra's algorithm \cite{dijkstra1959note}.

User-defined parameters can be used to customize number and width of the ligament landmarks.

\section{Experiments and Results}
\label{experiments}

To establish the reliability of the proposed method, we used the publicly available VerSe dataset \cite{Sekuboyina.2021}, the artificial Sawbone dataset \cite{wataname2018}, and a cervical spine dataset published in \cite{AlDhamari2019ADO}. In total, 57 three-dimensional vertebral surface models were augmented with manual annotations of the ligament attachment sites to use as ground truth. The annotations were derived based on anatomical knowledge and relevant literature and were acquired in collaboration with medical professionals.

\begin{figure}
    \centering
    \includegraphics[width=\linewidth, trim=6 1 1 1, clip]{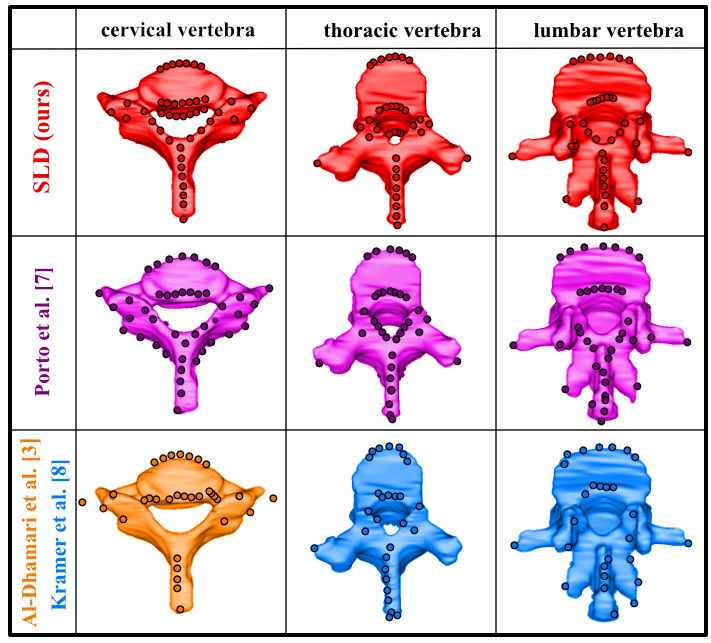}
    \caption{Comparison of four landmark detection methods (our proposed SLD method, \cite{AlDhamari2019ADO}, \cite{Porto2021ALP}, and \cite{Kramer.2024}), applied to cervical, thoracic and lumbar vertebra samples.}
    \label{fig:comparison-methods-landmarks}
\end{figure}

The visual results (see Fig.~\ref{fig:comparison-methods-landmarks}) indicate that the detected landmarks generally align with the intended ligament positions.
Minor errors were observed only for the ALL in cervical vertebrae, the ITL in thoracic vertebrae, and the CL in lumbar vertebrae.

We evaluated our method by comparing the detected landmarks to the annotated ground truth data.
Each landmark position $y_i$ was compared to the closest position $\hat{y}_i$ on the annotated ligament attachment site, and evaluated using the mean absolute error metric (MAE) and the root mean square error metric (RMSE).
Additionally, the standard deviation (SD) of the absolute errors, and the minimum and maximum errors (error range) were calculated. 
The evaluation shows high accuracy of the method, with an overall MAE of 0.7~mm and an RMSE of 1.1~mm. The results are displayed in Table~\ref{tab:landmarks_table_metrics}. Among the different spine regions, the lowest error values were observed for thoracic vertebrae (MAE~=~0.6~mm, RMSE~=~1.0~mm), while the highest error values were observed for cervical vertebrae (MAE~=~0.8~mm, RMSE~=~1.3~mm).
Among the different ligament groups, the lowest error values were observed for ALL (MAE~=~0.4~mm, RMSE~=~0.6~mm), while the highest error values were observed for ITL (MAE~=~1.5~mm, RMSE~=~2.3~mm).

\begin{table}[H]
    \centering
    \renewcommand{\arraystretch}{1.2}
    \begin{tabular}{|
    >{\columncolor[HTML]{EFEFEF}}l |c|c|c|}
    \hline
    \cellcolor[HTML]{C0C0C0}\textbf{Category} & \cellcolor[HTML]{C0C0C0}\textbf{MAE ± SD} & \cellcolor[HTML]{C0C0C0}\textbf{Error range} & \multicolumn{1}{r|}{\cellcolor[HTML]{C0C0C0}\textbf{RMSE}} \\ \hline
    \textit{\textbf{Overall}}        & \textbf{0.7 ± 0.9}                              & \textbf{0.0 – 9.6}                                & \textbf{1.1}                                                    \\ \hline
    \textit{\textbf{Cervical}}       & 0.8 ± 1.1                                       & 0.0 – 9.6                                         & 1.3                                                             \\
    \textit{\textbf{Thoracic}}       & 0.6 ± 0.8                                       & 0.0 – 7.7                                         & 1.0                                                             \\
    \textit{\textbf{Lumbar}}         & 0.8 ± 1.0                                       & 0.0 – 6.6                                         & 1.2                                                             \\ \hline
    \textit{\textbf{ALL}}            & 0.4 ± 0.5                                       & 0.0 – 5.5                                         & 0.6                                                             \\
    \textit{\textbf{PLL}}            & 0.5 ± 0.5                                       & 0.0 – 6.7                                         & 0.7                                                             \\
    \textit{\textbf{CL}}             & 1.0 ± 1.2                                       & 0.0 – 8.1                                         & 1.6                                                             \\
    \textit{\textbf{ISL}}            & 0.5 ± 0.7                                       & 0.0 – 7.7                                         & 0.9                                                             \\
    \textit{\textbf{LF}}             & 1.1 ± 0.9                                       & 0.0 – 6.1                                         & 1.5                                                             \\
    \textit{\textbf{ITL}}            & 1.5 ± 1.8                                       & 0.1 – 9.6                                         & 2.3                                                             \\
    \textit{\textbf{SSL}}            & 1.3 ± 1.1                                       & 0.1 – 5.9                                         & 1.7                                                             \\ \hline
    \end{tabular}
    \caption{Experimental results of our proposed SLD method. All values are reported in millimeters (mm).}
    \label{tab:landmarks_table_metrics}
\end{table}

Additionally, our proposed method for the segmentation-based landmark detection (SLD) was compared to three methods for landmark detection \cite{AlDhamari2019ADO,Porto2021ALP,Kramer.2024}. The experiments for the comparison methods were conducted seperately for different subsets of the datasets, since they are only applicable to specific input data such as vertebrae of specific spine regions. The results (see Table~\ref{tab:landmarks_table_metrics_comparison}) demonstrate that the proposed SLD method generally outperforms the comparison methods in terms of the MAE and the RMSE.

\begin{table}[htb]
    \centering
    \setlength{\tabcolsep}{3.7pt} 
    \renewcommand{\arraystretch}{1.3}
    \begin{tabular}{|
    >{\columncolor[HTML]{EFEFEF}} l | c |
    c|c|c|}
    \hline
    \cellcolor[HTML]{C0C0C0}\textbf{Method} & \cellcolor[HTML]{C0C0C0}\textbf{MAE ± SD} & \cellcolor[HTML]{C0C0C0}\textbf{Error range} & \multicolumn{1}{r|}{\cellcolor[HTML]{C0C0C0}\textbf{RMSE}} \\ \Xhline{2.5\arrayrulewidth}
    \textit{\textbf{SLD (Ours)}}          & \textbf{1.0 ± 1.3}             & \textbf{0.0 – 9.6}              & \textbf{1.6}         \\ \hline
    Al-Dhamari et al. \textit{\cite{AlDhamari2019ADO}}   & 2.6 ± 2.2                      & 0.1 – 15.8                      & 3.4                  \\ \Xhline{2.5\arrayrulewidth}
    \textit{\textbf{SLD (Ours)}}          & \textbf{0.7 ± 0.9}             & \textbf{0.0 – 9.6}              & \textbf{1.1}         \\ \hline
    Porto et al. \textit{\cite{Porto2021ALP}}       & 8.0 ± 12.9                     & 0.0 – 101.5                     & 15.2                 \\ \Xhline{2.5\arrayrulewidth}
    \textit{\textbf{SLD (Ours)}}          & \textbf{0.7 ± 0.8}             & \textbf{0.0 – 7.7}              & \textbf{1.1}         \\ \hline
    Kramer et al. \textit{\cite{Kramer.2024}}      & 3.4 ± 5.7                      & 0.0 – 53.0                      & 6.6                  \\ \hline 
    \end{tabular}
    \caption{Experimental results of the proposed SLD method, compared to three methods for landmark detection \cite{AlDhamari2019ADO,Porto2021ALP,Kramer.2024} for different subsets (row 1, cervical), (row 2, all regions) and (row 3, thoracic and lumbar) of the dataset. All values are reported in millimeters (mm).}
    \label{tab:landmarks_table_metrics_comparison}
\end{table}

\section{Discussion and Conclusion}
\label{conclusion}

In this work, we presented a novel segmentation-based method for the automated detection of spinal ligament landmarks, generalized across the cervical, thoracic, and lumbar regions. The proposed approach achieved high accuracy and outperformed existing automated methods in terms of MAE and RMSE, demonstrating strong generalization across spinal regions. Beyond its accuracy, the method proved robust across datasets of varying mesh quality and anatomical diversity, requiring no manual annotation or user interaction, unlike in the method of Al-Dhamari et al. \cite{AlDhamari2019ADO}. Its modular design enables seamless integration into existing segmentation or registration workflows, providing a reliable preprocessing step for the automated construction of biomechanical models. Registration-based state-of-the-art methods \cite{AlDhamari2019ADO,Porto2021ALP,Kramer.2024} tend to exhibit higher localization errors due to imperfect registration, whereas the proposed rule-based approach eliminates this dependency and achieves more robust performance.
Future work will address current limitations, including the treatment of highly variable ligaments such as the ITL and SSL, and the inclusion of upper cervical vertebrae (C1--C2). Integrating learning-based representations of vertebral geometry may further improve landmark accuracy and allow adaptation to pathological cases. 
Although the detection error of ligament attachment points on the vertebral surfaces is small, the impact of these errors on simulation outcomes remains unknown. Therefore, sensitivity studies will be performed in future.


The approach offers potential benefits for both clinical and research applications, including patient-specific finite element (FE) and multibody system (MBS) simulations, implant design, and functional spine analysis. Furthermore, the released dataset and implementation promote reproducibility and can serve as a foundation for future developments in computational spinal modeling.

\section{Compliance with ethical standards}
\label{sec:ethics}

This research study was conducted retrospectively using human subject data made available in open access. Ethical approval was not required as confirmed by the license attached with the open access data.

\bibliographystyle{IEEEbib}
\bibliography{strings,refs}

\end{document}